\documentclass{article} 
\usepackage{iclr2026_conference,times}

\usepackage[utf8]{inputenc}
\usepackage[T1]{fontenc}

\usepackage{amsmath}
\usepackage{amsfonts}
\usepackage{graphicx}
\usepackage{wrapfig}
\usepackage{booktabs}
\usepackage{adjustbox}
\usepackage{nicefrac}
\usepackage{microtype}
\usepackage{pifont}
\usepackage[table]{xcolor}
\usepackage{hyperref}
\usepackage{url}
\usepackage{fancyhdr}

\renewcommand{\headrulewidth}{0pt}
\renewcommand{\footrulewidth}{0pt}

\fancypagestyle{plain}{
    \fancyhf{}
    \fancyfoot[C]{\thepage}
    \renewcommand{\headrulewidth}{0pt}
    \renewcommand{\footrulewidth}{0pt}
}

\definecolor{myblue}{rgb}{0.88,0.98,1}
\definecolor{mygreen}{rgb}{0.92, 1.0, 0.92}
\definecolor{myred}{rgb}{1, 0.9, 0.9}
\usepackage[most]{tcolorbox}

\definecolor{promptbg}{RGB}{248,248,248}
\definecolor{promptframe}{RGB}{180,180,180}

\newtcolorbox{promptbox}[1][]{
    enhanced,
    breakable,
    colback=promptbg,
    colframe=promptframe,
    boxrule=0.5pt,
    arc=2pt,
    left=6pt,
    right=6pt,
    top=6pt,
    bottom=6pt,
    fonttitle=\bfseries,
    title=#1
}

\iclrfinalcopy

\usepackage[utf8]{inputenc} 
\usepackage[T1]{fontenc}    
\usepackage{url}            
\usepackage{booktabs}       
\usepackage{amsfonts}       
\usepackage{nicefrac}       
\usepackage{microtype}      
\usepackage{xcolor}         
\usepackage{pifont}
\usepackage[table]{xcolor}
\usepackage{graphicx}
\usepackage{adjustbox}
\definecolor{myblue}{rgb}{0.88,0.98,1}
\definecolor{mygreen}{rgb}{0.92, 1.0, 0.92}
\definecolor{myred}{rgb}{1, 0.9, 0.9}

\definecolor{cityblue}{RGB}{0,114,178}
\hypersetup{
    colorlinks=true,
    urlcolor=cityblue,
    linkcolor=cityblue,
    citecolor=cityblue
}
\newcommand{\projecturl}{https://lijayutnt.github.io/ShutterMuse/}
\newcommand{\benchmarkurl}{https://huggingface.co/datasets/ShutterMuse/CaptureGuide-Bench}
\newcommand{\modelurl}{https://huggingface.co/ShutterMuse/ShutterMuse}
\newcommand{\codeurl}{https://github.com/lijayuTnT/ShutterMuse}
\newcommand{\linkicon}[3]{%
    \href{#1}{%
        \raisebox{-0.2\height}{\includegraphics[height=0.45cm]{#2}}~
        \textcolor{cityblue}{\textbf{#3}}%
    }%
}
\title{
\texorpdfstring{
\makebox[\textwidth][c]{%
\begin{tabular}{c}
\makebox[0pt][r]{%
\raisebox{-0.22\height}{\includegraphics[height=0.95cm]{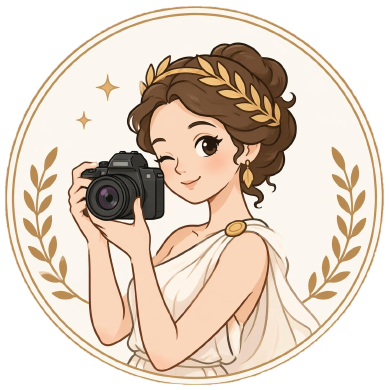}}
}%
ShutterMuse: Capture-Time\\
Photography Guidance with MLLMs
\end{tabular}%
}
}{ShutterMuse: Capture-Time Photography Guidance with MLLMs}
}

\author{
\begin{tabular}{c}
Jiayu Li$^{1,2}$ \quad
Yixiao Fang$^{2,\dagger}$ \quad
Tianyu Hu$^{2}$ \quad
Wei Cheng$^{2}$ \quad
Ping Huang$^{2}$ \\
Zheheng Fan$^{2}$ \quad
Gang Yu$^{2,\ddagger}$ \quad
Xingjun Ma$^{1,\ddagger}$ \\[0.4em]
$^{1}$Fudan University \quad
$^{2}$StepFun \\[0.4em]
\normalsize
\linkicon{\projecturl}{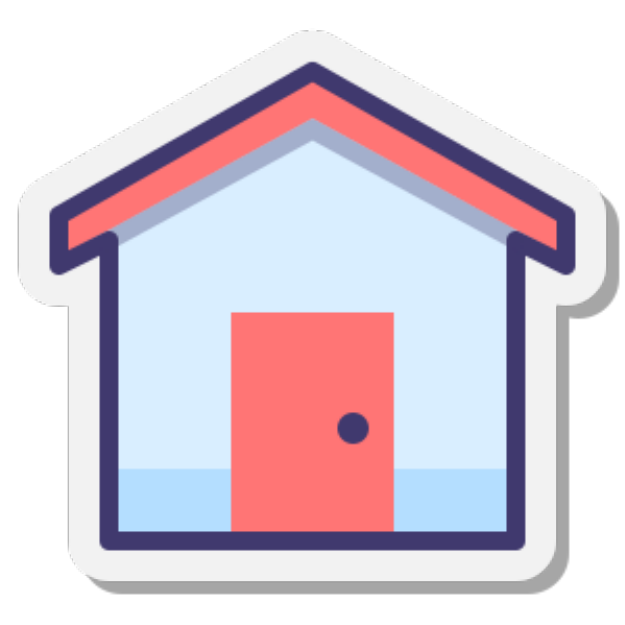}{Project Page}
\quad
\linkicon{\benchmarkurl}{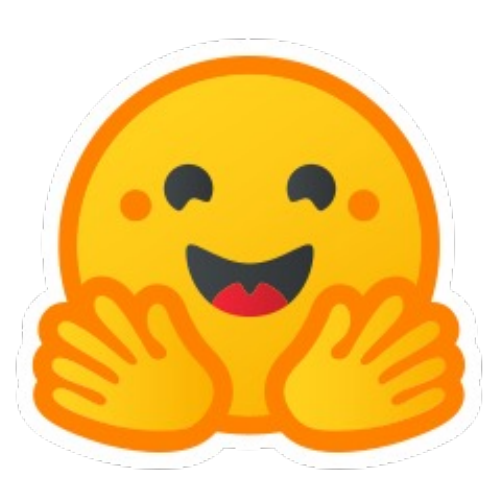}{Benchmark}
\quad
\linkicon{\modelurl}{Fig/huggingface_logo.pdf}{Models}
\quad
\linkicon{\codeurl}{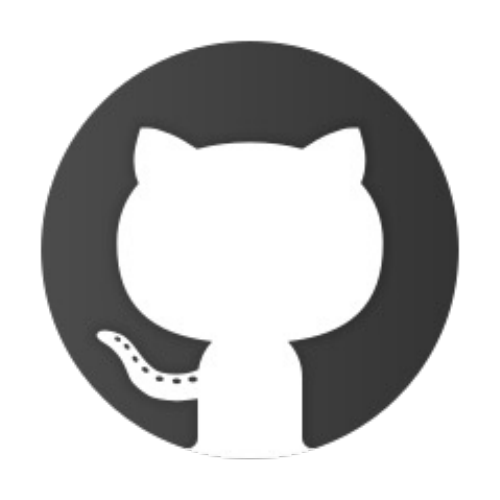}{Code}
\end{tabular}
}

\begin{document}

\maketitle
\thispagestyle{plain}
\pagestyle{plain}

\begin{figure}[h]
    \centering
    \vspace{-25pt}
    \includegraphics[width=\linewidth]{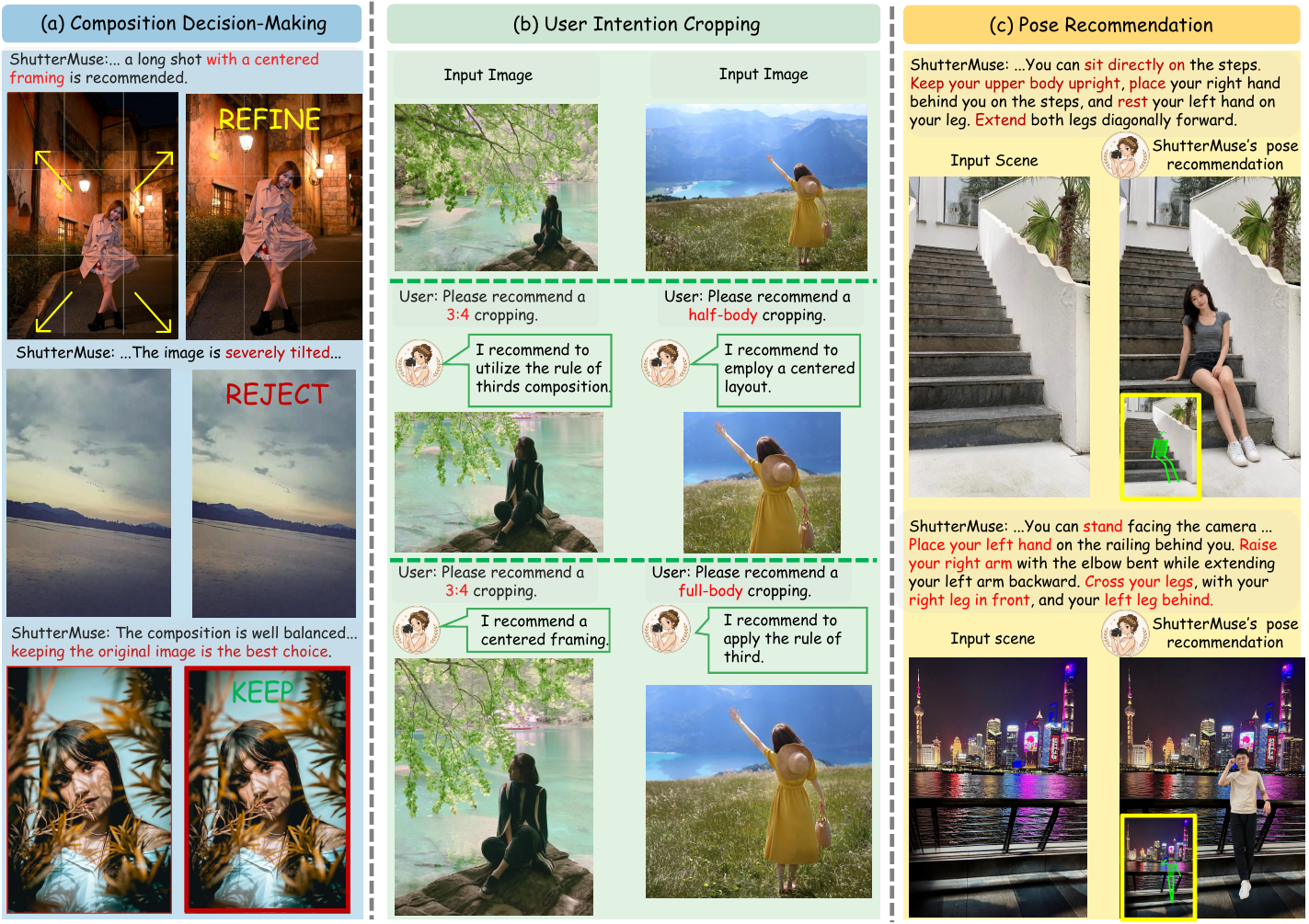}
    \vspace{-20pt}
    \caption{
    \textbf{Showcases of ShutterMuse.}
    ShutterMuse supports both photographer-side and subject-side guidance.
    (a) ShutterMuse can determine whether and how to adjust the composition. (b) ShutterMuse can understand and respond to diverse user intentions. (c) ShutterMuse can provide scene-conditioned pose recommendation. The keypoint poses are rendered by GPT-Image-2.
    }
    \label{fig:teaser}
\end{figure}

{\let\thefootnote\relax\footnotetext{\noindent$\dag$ Yixiao Fang leads this project; \ddag Corresponding authors.}}

\begin{abstract}
Real-world photography requires capture-time guidance for both camera framing and subject pose. Yet existing aesthetic cropping benchmarks mainly evaluate post-hoc crop prediction and overlook subject-side recommendations, leaving the capture-time guidance capabilities of multimodal large language models (MLLMs) underexplored. To address this gap, we introduce \textbf{CaptureGuide-Bench}, a benchmark with two complementary tasks: photographer-side composition decision and refinement, and subject-side scene-conditioned pose recommendation. Our evaluation reveals limitations: general-purpose MLLMs can make composition decisions but lack precise refinement localization, while specialized aesthetic cropping models localize crops effectively but are limited to refinement; neither provides actionable pose guidance. To support model development, we further construct \textbf{CaptureGuide-Dataset}, comprising 130K samples with textual rationales and structured visual annotations, and develop \textbf{ShutterMuse}, a unified MLLM trained with supervised and reinforcement fine-tuning. Experiments on CaptureGuide-Bench show that ShutterMuse achieves the best overall photographer-side performance among evaluated baselines and competitive subject-side pose recommendation with substantially lower inference cost, demonstrating the potential of MLLMs as interactive assistants for photography during image capture.
\end{abstract}

\section{Introduction}
\label{sec:intro}

Recent advances in multimodal large language models (MLLMs) have improved visual understanding, aesthetic reasoning, and instruction following \citep{alayrac2022flamingo,li2023blip,dai2023instructblip,zhu2024minigpt,bai2025qwen3,bai2023qwenvlversatilevisionlanguagemodel,liu2023visual,team2023gemini,wang2025internvl3,huang2024aesexpert,qi2025photographer,wu2023q,liu2025advancing,cao2025artimuse}. However, their capability to provide photography guidance during image capture remains underexplored. In real-world photography, the photographer needs to decide whether the current framing should be kept, refined, or rejected, while the subject may need pose guidance that better matches the scene. Existing aesthetic cropping benchmarks \citep{yan2013learning,hong2024learning,yang2023focusing,zeng2019reliable,wei2018good,chen2017quantitative,fang2014automatic,zhang2022human} mainly formulate photography guidance as post-hoc crop prediction and typically assume that each image can be improved by cropping. As a result, they cannot fully evaluate capture-time photography guidance, especially when cropping is unnecessary, insufficient, or when subject-side guidance is required.

\begin{figure}[h]
    \centering
    \includegraphics[width=1.0\linewidth]{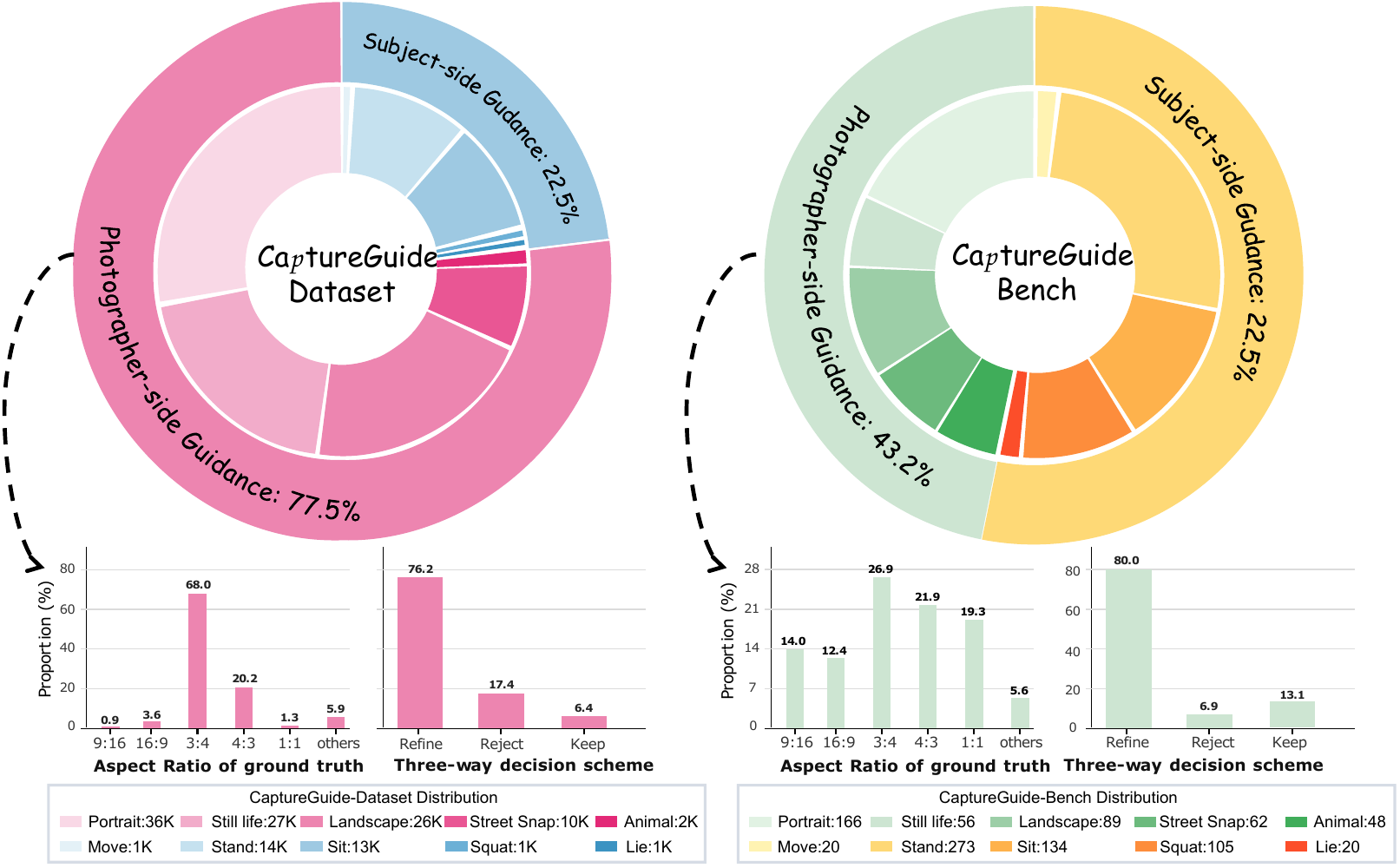}
    \caption{Distribution of our dataset and benchmark.}
    \label{data_distribution}
\end{figure}

To address this gap, we introduce \textbf{CaptureGuide-Bench}, a benchmark for evaluating capture-time photography guidance with MLLMs. As shown in Fig.~\ref{data_distribution}, it covers two complementary tasks: \textbf{photographer-side guidance} and \textbf{subject-side guidance}. Photographer-side guidance, which spans five representative photography scenarios, evaluates whether a model can make a three-way decision among \texttt{refine}, \texttt{keep}, and \texttt{reject}, and produce a valid framing box when refinement is needed. Subject-side guidance covers five common human poses and evaluates whether a model can recommend a scene-conditioned human pose. Our evaluation reveals complementary limitations of existing models: general-purpose MLLMs \citep{wang2025internvl3,bai2025qwen3,team2023gemini,achiam2023gpt,moonshotai2025kimik26} can often make composition decisions but lack precise refinement localization, whereas specialized aesthetic cropping models \citep{sheng2025instructcrop,du2026venus,hong2021composing,liu2023beyond} localize crops effectively but are limited to refinement and cannot handle \texttt{keep} or \texttt{reject}; neither supports structured, actionable subject-side guidance.

To support model development, we construct \textbf{CaptureGuide-Dataset}, a large-scale dataset with approximately 130K samples. Its photographer-side subset covers five representative photography scenarios and six common composition aspect ratios, with composition annotations scaled through an expert-seeded, MLLM-verified self-distillation pipeline. Its subject-side subset covers five common human pose types and is built by converting portrait images into person-free scenes paired with expert-verified pose keypoints, visibility states, and rationales. Based on this dataset, we propose \textbf{ShutterMuse}, a unified MLLM trained with supervised fine-tuning and reinforcement fine-tuning for structured capture-time guidance.

Extensive experiments on CaptureGuide-Bench show that ShutterMuse achieves the best overall performance among evaluated baselines on photographer-side guidance, while producing competitive subject-side pose recommendations with substantially lower inference cost. These results demonstrate the potential of MLLMs as interactive assistants for photography during image capture.

Our main contributions are summarized as follows:
\begin{itemize}
    \item We introduce \textbf{CaptureGuide-Bench}, a benchmark for evaluating \emph{capture-time photography guidance}, covering both photographer-side composition decision, refinement and subject-side pose recommendation.
    
    \item We construct \textbf{CaptureGuide-Dataset}, a large-scale dataset with approximately 130K samples, including textual rationales, structured composition boxes, pose keypoints, and visibility states.
    
    \item We propose \textbf{ShutterMuse}, a unified MLLM trained with supervised fine-tuning and reinforcement fine-tuning to generate structured and interpretable capture-time guidance.
    
    \item Experiments show that ShutterMuse achieves the best overall photographer-side performance among evaluated baselines and provides efficient, competitive subject-side pose recommendations.
\end{itemize}

\section{Related Work}

\subsection{Aesthetic Image Cropping Benchmarks and Datasets}

Aesthetic image cropping has traditionally been formulated as a \emph{post-hoc} composition refinement problem \citep{yan2013learning,hong2024learning,yang2023focusing,zeng2019reliable,wei2018good,chen2017quantitative,fang2014automatic,zhang2022human}, where a captured image is improved by predicting a better crop. Early benchmarks such as FCDB \citep{chen2017quantitative} and FLMS \citep{fang2014automatic} established this setting by collecting expert-annotated cropping results. More recently, SACD \citep{zhang2022human} further advanced this direction by introducing subject-aware annotations, enabling large-scale learning of cropping preferences. However, these benchmarks are still centered on photographer-side post-processing and generally assume that each image admits a preferable crop.

\subsection{Aesthetic Cropping and Composition Recommendation}

Existing aesthetic cropping methods can be broadly divided into proposal-based methods \citep{su2024spatial,zeng2019reliable,zhang2026procrop} and regression-based methods \citep{hong2021composing,huang2024multi,pan2021robust,du2026venus,sheng2025instructcrop}. Proposal-based methods rank candidate crops generated from anchor boxes or sliding windows using aesthetic cues such as saliency and photographic rules \citep{ni2013learning,fang2014automatic,zeng2019reliable}. Regression-based methods, in contrast, directly predict crop coordinates in an end-to-end manner. Recent MLLM-based approaches have further extended this paradigm by introducing instruction following, explanation generation, and aesthetic reasoning. Specifically, InstructCrop \citep{sheng2025instructcrop} constructs an instruction-tuning dataset, enabling MLLMs to generate explanatory and intention-aware crop suggestions. Venus \citep{du2026venus} equips MLLMs with aesthetic guidance capabilities through a two-stage training strategy. Nevertheless, these methods remain primarily designed for after-capture, photographer-side crop adjustment rather than interactive guidance during image capture.

\subsection{Human Pose and Motion Generation}

Recent human motion generation methods have demonstrated strong capabilities in synthesizing controllable body motions from language instructions \citep{tevet2022human,chen2023executing,zhang2023generating,jiang2023motiongpt,guo2024momask,zhang2023finemogen}. T2M-GPT \citep{zhang2023generating} and MotionGPT \citep{jiang2023motiongpt} model motion as discrete token sequences for text-driven generation, while MoMask \citep{guo2024momask} improves text-to-motion synthesis through hierarchical tokenization and masked generative modeling. However, these methods mainly address generic text-conditioned motion synthesis, completion, or editing, rather than scene-grounded pose recommendation. In contrast, our method generates scene-conditioned poses for capture-time photography guidance.

\section{Dataset and Benchmark}
We present \textbf{CaptureGuide-Dataset} as illustrated in Fig.~\ref{data_distribution}, a large-scale dataset for real-world capture-time photography guidance. It contains approximately 130K images in total, including 100K photographer-side guidance samples and 30K subject-side guidance samples. To facilitate standardized evaluation and downstream method development, we further introduce \textbf{CaptureGuide-Bench}, a benchmark built on top of the dataset for both guidance tasks.

\begin{figure}[h]
    \centering
    \includegraphics[width=1.0\linewidth]{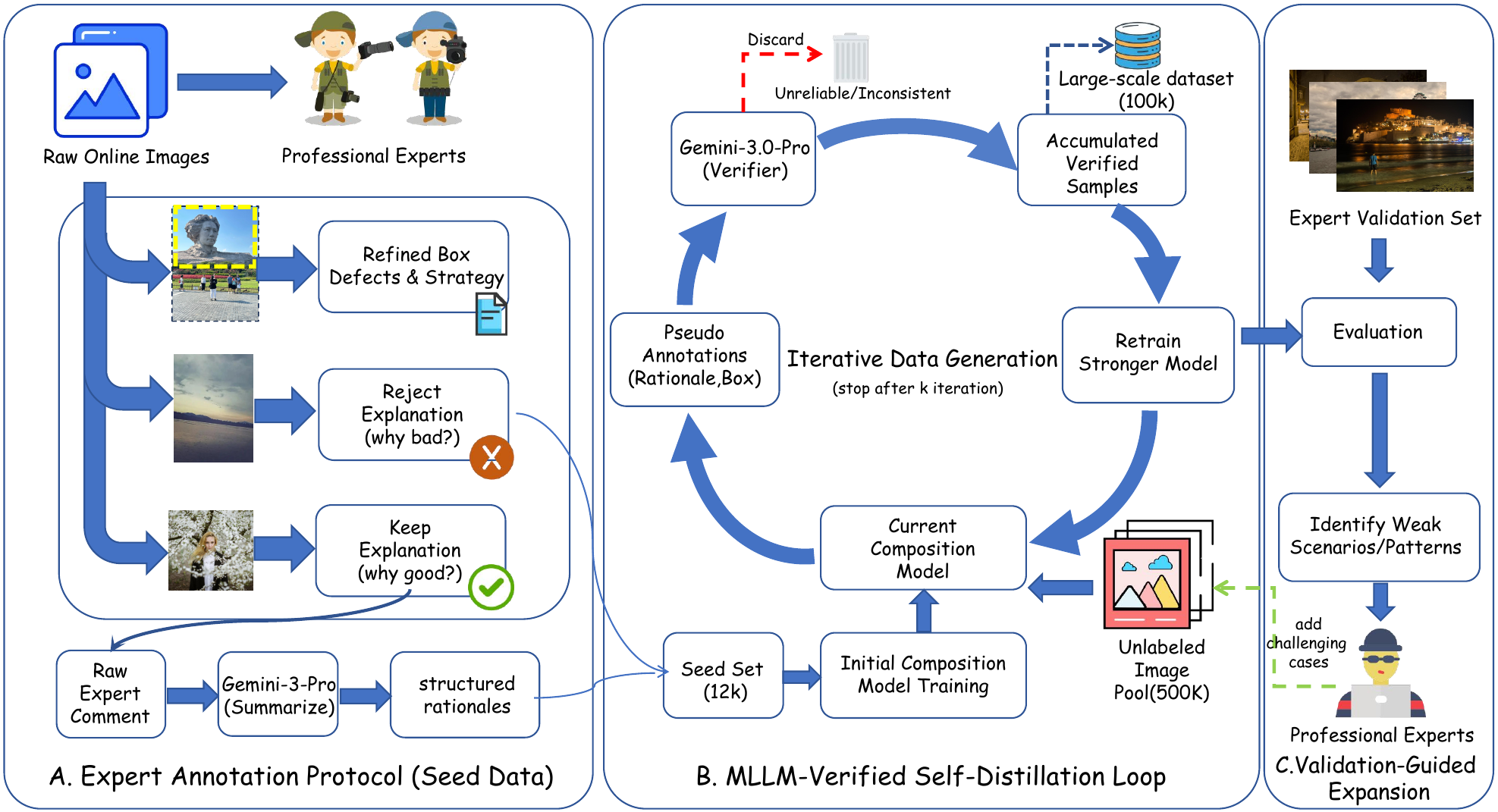}
    \caption{
Overview of the expert-seeded, MLLM-verified self-distillation pipeline for photographer-side data construction. Expert seed annotations are structured by an MLLM, expanded through pseudo-labeling and verification, and monitored with a held-out validation set.
}
    \label{fig:composition_data_construction}
\end{figure}

\begin{figure}[h]
    \centering
    \includegraphics[width=1.0\linewidth]{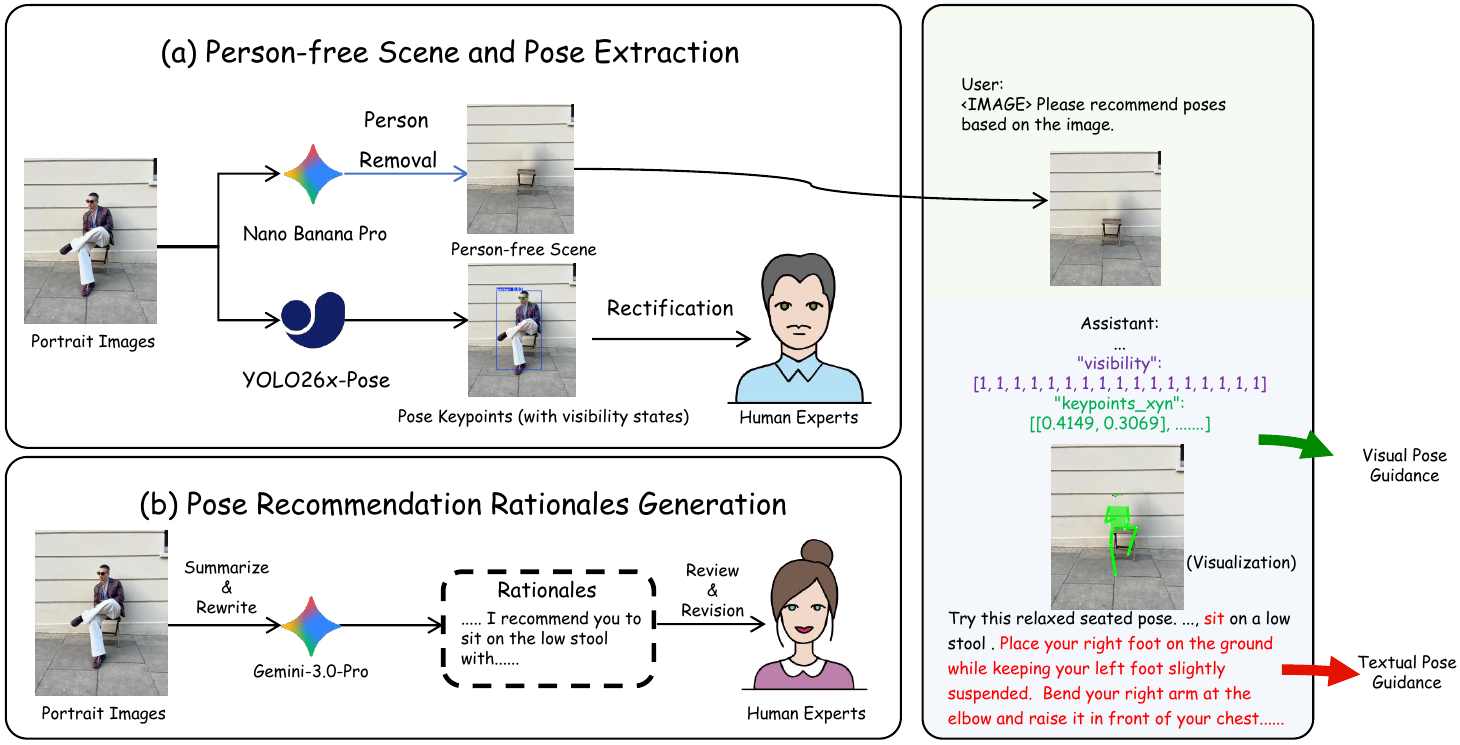}
\caption{
Overview of the subject-side guidance generation pipeline. Portrait images are converted into person-free scenes and paired with verified pose keypoints, visibility states, and textual rationales.
}
    \label{fig:pose_data_construction}
\end{figure}

\subsection{CaptureGuide-Dataset}
\paragraph{Photographer-side guidance.}
As illustrated in Fig.~\ref{fig:composition_data_construction}, we first construct an expert-labeled seed set for photographer-side guidance. Images are collected from multiple online platforms and annotated with one of three decisions: \texttt{refine}, \texttt{reject}, or \texttt{keep}. For \texttt{refine} samples, annotators provide a refined composition box and free-form comments describing the composition defects and the recommended reframing strategy. For \texttt{keep} samples, annotators explain the compositional strengths of the original image, while for \texttt{reject} samples, they describe non-croppable defects that make the image unsuitable for recommendation. Moreover, we use an MLLM to summarize and normalize the raw expert comments into structured rationales, which are paired with the corresponding decision labels and composition boxes when applicable. Detailed annotation guidelines for the three categories are provided in Appendix~\ref{app:photographer_guidelines}.

To ensure annotation quality, the seed set is labeled by 10 trained annotators with cross-review, and ambiguous or low-agreement cases are re-annotated. This process yields a high-quality seed set of 12K images. Although this seed set provides reliable supervision for photographer-side guidance, scaling expert annotation is costly because it requires both aesthetic judgment and feasible reframing decisions. To scale the annotations, we adopt an \textit{expert-seeded, MLLM-verified self-distillation pipeline} (EMDP), as shown in Fig.~\ref{fig:composition_data_construction}. Starting from the expert seed set, we train an initial composition model, use it to generate pseudo annotations on unlabeled images, and filter them with an MLLM verifier that checks rationale correctness and rationale-box consistency. Verified samples are then used for iterative retraining. To reduce error accumulation, we maintain a fixed expert validation set for monitoring weak composition patterns and reserve an independent expert test set for evaluating pipeline reliability.

\paragraph{Subject-side guidance.}
To support subject-side guidance, each training sample is formulated as a triplet consisting of a person-free scene image, a target human pose represented by keypoints, and textual rationales explaining why the pose suits the scene. As illustrated in Fig.~\ref{fig:pose_data_construction}, we construct these triplets through a \textit{subject-side guidance generation pipeline} (SGGP).

Given a portrait image, we first remove the person using Nano-Banana-Pro \cite{google_nanobananapro} to obtain an empty scene while preserving the background layout and context.
In parallel, we extract initial human keypoints from the original portrait image using a YOLO-based pose estimator \citep{jocher2026ultralyticsyolo26unifiedrealtime} in the standard COCO 17-keypoint format \cite{lin2014microsoft}.
We then prompt Gemini-3.0-Pro to analyze the original portrait image, summarize the scene context and human pose, and rewrite the analysis into pose-recommendation rationales that explain why the observed pose is suitable for the scene. Professional annotators subsequently review and revise these rationales to ensure that they are accurate, contextually grounded, and expressed in a clear recommendation-oriented style.

To handle common occlusion and truncation in portrait photography, each keypoint is assigned one of three visibility states: visible in the image, invisible but within the image, or outside the image frame. The detailed COCO-17 keypoint order, visibility definitions, and human verification procedure are described in Appendix~\ref{app:subject_guidelines}. We first apply confidence-based filtering to remove unreliable keypoint predictions and then ask annotators to correct inaccurate or missing keypoints. Five experienced photographers further verify both the generated rationales and the pose annotations, checking whether the rationales are consistent with the scene context and whether the keypoints and visibility states accurately describe the human pose. This process yields 30K person-free scene images paired with expert-verified pose keypoints and structured rationales.

\subsection{CaptureGuide-Bench}
\label{captureguide-bench}
To address the lack of standardized evaluation protocols for capture-time photography guidance, we introduce \textbf{CaptureGuide-Bench}, a benchmark covering both photographer-side and subject-side guidance. The benchmark consists of two complementary subsets: one for composition decision-making and one for subject-side pose guidance.

For photographer-side guidance, we follow the same expert seed construction protocol as in EMDP
and collect 421 held-out samples covering the three-way decision scheme and diverse photographic subjects. For \textit{refine} samples, we annotate 3--5 ground-truth bounding boxes per image.

For subject-side guidance, we sample 552 examples from the generated subject-side dataset with balanced coverage of pose types and scene types. All samples in CaptureGuide-Bench are held out from model training and are not used in either the SFT or RFT stages.

\paragraph{Photographer-side evaluation metrics.}
For photographer-side guidance, the model predicts a decision $d \in \{\texttt{refine}, \texttt{keep}, \texttt{reject}\}$ for each input image. If $d=\texttt{refine}$, the model additionally outputs a framing box $b=(x_1,y_1,x_2,y_2)$. Following prior works \citep{fang2014automatic,zhang2022human}, for each \textit{refine} sample, we compute Intersection-over-Union (IoU) as the maximum overlap between the predicted box and all annotated boxes, and the minimum boundary displacement error (BDE) over the annotated boxes. We also report the refinement success rate (R), defined as the percentage of samples with IoU larger than 0.7. 
For non-refinement decisions, we report the reject success rate (RSR) and keep success rate (KSR), which are defined as the percentages of ground-truth reject and keep samples that are correctly classified, respectively.

To assess compositional quality beyond geometric overlap, we further report MLLM-Score. Since
photographer-side guidance involves three possible decisions, we compute MLLM-Score in a
task-aware manner. For non-reject predictions, the predicted composition frame is used for evaluation:
a refinement box for \texttt{refine} predictions and the full image for \texttt{keep}
predictions. The MLLM judge then scores whether the resulting composition preserves the annotated
strengths and reasonably addresses the annotated weaknesses. If a model predicts \texttt{reject} for
a non-reject sample or outputs an invalid frame, we assign a score of 0. For ground-truth
\texttt{reject} samples, where no valid composition is expected, we assign a score of 1 if the model correctly predicts \texttt{reject}, and 0 otherwise. The final
MLLM-Score is averaged over the photographer-side benchmark. The judging prompt for non-reject compositions is provided in Appendix~\ref{app:photographer_mllm_score_prompt}.

\paragraph{Subject-side evaluation metrics.}
Pose recommendation cannot be adequately evaluated with a single geometric criterion, since multiple poses may be plausible for the same scene. Accordingly, the reference keypoints in CaptureGuide-Bench are used to characterize plausible pose configurations rather than as the sole target for exact geometric matching. In practice, users care more about whether the pose is physically plausible, semantically aligned with the scene, and visually appealing. Therefore, we render the predicted keypoints as a skeleton overlay on the input scene and use an MLLM to assess pose quality along three dimensions: \textbf{physical plausibility}, \textbf{scene interaction}, and \textbf{pose aesthetics}. The prompt templates for all three dimensions are provided in Appendix~\ref{app:subject_mllm_score_prompt}.
\section{ShutterMuse}

\textbf{Framework.}
We propose \textbf{\textit{ShutterMuse}}, a multimodal large language model (MLLM) built on Qwen3-VL-8B \cite{bai2025qwen3} for capture-time photography guidance. ShutterMuse supports both photographer-side guidance and subject-side guidance within a unified multimodal framework.

\textbf{Supervised Fine-Tuning.}
In the first stage, we perform supervised fine-tuning (SFT) on \textbf{\textit{CaptureGuide-Dataset}} to learn structured photography guidance generation. Each training sample consists of an input image $\mathbf{x}$, a text prompt $\mathbf{p}$, and a structured target response $\mathbf{y}$. The model is trained to follow the prompt and generate a JSON-formatted output, where the response schema depends on the guidance type. Specifically, for photographer-side guidance, the JSON response contains three fields: \texttt{task\_type}, \texttt{reason}, and \texttt{composition\_xy}. The field \texttt{task\_type} is set to \texttt{composition}. The field \texttt{composition\_xy} encodes the composition decision in a structured form: an empty value indicates \texttt{reject}, $[0,0,1,1]$ indicates \texttt{keep}, and $[x_1,y_1,x_2,y_2]$ indicates \texttt{refine}, where $(x_1,y_1,x_2,y_2)\in[0,1]^4$ and $[x_1,y_1,x_2,y_2]\neq[0,0,1,1]$. For subject-side guidance, the JSON response contains the fields \texttt{task\_type}, \texttt{reason}, \texttt{keypoints\_xyn}, and \texttt{visibility}. The field \texttt{task\_type} is set to \texttt{pose}. The field \texttt{keypoints\_xyn} stores the normalized coordinates of 17 human keypoints in the standard COCO 17-keypoint format. The field \texttt{visibility} is a 17-dimensional vector, where $1$ denotes a visible keypoint, $0$ denotes an occluded but within-image keypoint, and $-1$ denotes a keypoint outside the image frame.

Let $q=(\mathbf{x},\mathbf{p})$ denote the image-prompt input and
$\mathbf{y}^{\star}=(y^{\star}_1,\ldots,y^{\star}_L)$ denote the target JSON response.
We optimize the response-only next-token prediction loss:
\begin{equation}
\mathcal{L}_{\mathrm{SFT}}(\theta)
=
-\mathbb{E}_{(q,\mathbf{y}^{\star})\sim\mathcal{D}_{\mathrm{SFT}}}
\left[
\frac{1}{L}
\sum_{t=1}^{L}
\log
\pi_{\theta}
\left(
y^{\star}_t
\mid q,y^{\star}_{<t}
\right)
\right].
\end{equation}

\textbf{Reinforcement Fine-Tuning.}
In the second stage, we apply Group Relative Policy Optimization (GRPO) to further improve the model's decision-making ability and output accuracy. We construct a reinforcement learning dataset containing 20,000 samples following EMDP and SGGP. Given an image-prompt pair $(\mathbf{x}, \mathbf{p})$, the model generates responses and receives task-specific rewards according to the guidance type specified by \texttt{task\_type}.

For \textbf{photographer-side guidance}, we define two reward terms. Let $c^\star \in \{\texttt{reject}, \texttt{keep}, \texttt{refine}\}$ denote the ground-truth decision category, and let $\hat{c}$ be the predicted category parsed from \texttt{composition\_xy} using the same rule as in the SFT stage. The first reward measures whether the model predicts the correct three-way decision:
\begin{equation}
R_{\text{dec}} =
\begin{cases}
1, & \text{if } \hat{c} = c^\star, \\
0, & \text{otherwise}.
\end{cases}
\end{equation}
The second reward evaluates whether the model preserves the main subject when generating a refined composition box. Specifically, for samples whose ground-truth decision is \texttt{refine}, we use BiRefNet \citep{zheng2024bilateral}, an off-the-shelf salient object detection model, to extract a binary salient-object mask $M \in \{0,1\}^{H \times W}$ from the input image. Let $b$ denote the predicted composition box, and let $\mathbf{1}_{b}(u,v)$ indicate whether pixel $(u,v)$ lies inside $b$. We measure the mask coverage by
\begin{equation}
\operatorname{Cov}(b,M) =
\frac{\sum_{u,v} M(u,v)\mathbf{1}_{b}(u,v)}
{\sum_{u,v} M(u,v) + \epsilon}.
\end{equation}
The subject-preservation reward is defined as
\begin{equation}
R_{\text{mask}} =
\begin{cases}
1, & \text{if } c^\star = \texttt{refine} \text{ and } \operatorname{Cov}(b,M) \ge \tau_m, \\
0, & \text{otherwise},
\end{cases}
\end{equation}
where $\tau_m$ is a coverage threshold. The photographer-side reward is then given by
\begin{equation}
R_{\text{photo}} = R_{\text{dec}} + R_{\text{mask}}.
\end{equation}

For \textbf{subject-side guidance}, we use a single reward based on the visibility annotation. Let $\mathbf{v}_{\text{gt}} \in \{-1,0,1\}^{17}$ and $\mathbf{v}_{\text{pred}} \in \{-1,0,1\}^{17}$ denote the ground-truth and predicted visibility vectors, respectively. The reward is defined as
\begin{equation}
R_{\text{sub}} =
\begin{cases}
1, & \text{if } \mathbf{v}_{\text{pred}} = \mathbf{v}_{\text{gt}}, \\
0, & \text{otherwise}.
\end{cases}
\end{equation}

For each input $q$, we sample a group of $G$ responses
$\{\mathbf{y}_i\}_{i=1}^{G}$ from the old policy $\pi_{\theta_{\mathrm{old}}}$ and compute their
task-specific rewards $\{r_i\}_{i=1}^{G}$. The group-relative advantage is defined as
\begin{equation}
A_i =
\frac{r_i-\operatorname{mean}(\{r_j\}_{j=1}^{G})}
{\operatorname{std}(\{r_j\}_{j=1}^{G})+\epsilon}.
\end{equation}
For each token $y_{i,t}$, let
\begin{equation}
\rho_{i,t}(\theta)
=
\frac{
\pi_{\theta}(y_{i,t}\mid q,y_{i,<t})
}{
\pi_{\theta_{\mathrm{old}}}(y_{i,t}\mid q,y_{i,<t})
}.
\end{equation}
The GRPO loss is
\begin{equation}
\mathcal{L}_{\mathrm{GRPO}}(\theta)
=
-
\mathbb{E}
\left[
\frac{1}{G}
\sum_{i=1}^{G}
\frac{1}{L_i}
\sum_{t=1}^{L_i}
\left(
\min
\left(
\rho_{i,t} A_i,
\operatorname{clip}(\rho_{i,t},1-\epsilon_c,1+\epsilon_c) A_i
\right)
-
\beta
D_{\mathrm{KL}}
\left(
\pi_{\theta}
\|
\pi_{\mathrm{ref}}
\right)
\right)
\right],
\end{equation}
where the KL term is computed at each decoding step with the same conditioning context.

\section{Experiments}
\label{sec:experiments}
\subsection{Experimental Setup}
\paragraph{Implementation Details.}
Our ShutterMuse model is initialized from Qwen3-VL-8B and trained in two stages.
In the first stage, we perform supervised fine-tuning (SFT) on CaptureGuide-Dataset, constructed through our EMDP and SGGP pipelines. The EMDP pipeline is conducted for three rounds. SFT is performed on eight A800 GPUs using the AdamW optimizer with a learning rate of $1\times10^{-4}$ and an effective batch size of 64 for 5 epochs. During inference, we employ vLLM~\citep{kwon2023efficient} to accelerate decoding and improve inference throughput. In the second stage, we perform reinforcement fine-tuning (RFT) using the GRPO algorithm on a dedicated dataset of 20K samples. The RFT stage is initialized from the SFT model and uses the SFT model as the reference policy. For GRPO, we use an effective batch size of 64 and sample 32 rollouts per input for group-wise reward normalization and policy optimization. We train for 1 epoch with a learning rate of $1\times10^{-6}$, a weight decay of 0.1, and a KL regularization coefficient of $\beta=0.01$. For the mask-coverage reward used in photographer-side refinement, we set the coverage threshold to $\tau_m=0.9$.
\subsection{Evaluation Metrics}
We evaluate all methods on \textbf{CaptureGuide-Bench} using the metric definitions introduced in Sec.~\ref{captureguide-bench}. For \textbf{photographer-side guidance}, we report IoU and BDE to measure crop localization quality, and refinement success rate (R), reject success rate (RSR), and keep success rate (KSR) to evaluate decision-making performance. In addition, we report MLLM-Score to complement geometric measures such as IoU and BDE. For \textbf{subject-side guidance}, we report the average score across three criteria: physical plausibility, scene interaction, and pose aesthetics. We also measure efficiency in terms of the average number of generated tokens and inference time per recommendation.

For all MLLM-based evaluations, we use \textbf{Gemini-3.0-Pro} as the judge. Following the three-level scoring protocol described in Sec.~\ref{captureguide-bench}, each sample is assigned a score of $\{0, 0.5, 1\}$.

\subsection{Quantitative Analysis}

\begin{table}[h]
  \centering
  \caption{Quantitative results on the photographer-side guidance subset of our CaptureGuide-Bench. \textbf{RSR} and \textbf{KSR} denote reject success rate and keep success rate, respectively. Best and second-best results are in \textbf{bold} and \underline{underlined}, respectively. Tied best results are all bolded.}
  \label{tab:photographer_side_bench}
  \begin{adjustbox}{max width=\linewidth}
  \small
  \renewcommand{\arraystretch}{1}
  \setlength{\tabcolsep}{1.1mm}
  \begin{tabular}{l|cccccc}
    \toprule
    \textbf{Method} & IoU\%($\uparrow$) & BDE($\downarrow$) & R\%($\uparrow$) & RSR\%($\uparrow$) & KSR\%($\uparrow$) & MLLM-Score($\uparrow$) \\
    \midrule
    \rowcolor{myred}\multicolumn{7}{c}{\textbf{Open-source General MLLMs}} \\
    \midrule
    InternVL3.5-8B~\cite{wang2025internvl3} & 42.86 & 0.127 & 8.61 & 0.00 & 20.00 & 0.15 \\
    Kimi-K2.6~\cite{moonshotai2025kimik26} & 65.44 & 0.087 & 37.92 & 0.00 & 90.90 & 0.47 \\
    Qwen3-VL-8B-Instruct~\cite{bai2025qwen3} & 55.18 & 0.105 & 18.40 & 0.00 & 36.36 & 0.25 \\
    Qwen3-VL-32B-Instruct~\cite{bai2025qwen3} & 63.80 & 0.101 & 35.91 & 13.79 & \textbf{98.18} & 0.47 \\
    Qwen3-VL-235B-A22B-Instruct~\cite{bai2025qwen3} & 61.84 & 0.093 & 33.53 & 20.69 & \underline{94.55} & 0.48 \\
    Qwen3.5-9B~\cite{qwen35blog} & 61.94 & 0.094 & 30.86 & 3.45 & 83.64 & 0.45 \\
    Qwen3.6-27B~\cite{qwen36_27b} & 53.93 & 0.090 & 33.23 & 48.28 & 72.72 & 0.47 \\ 
    \midrule
    \rowcolor{mygreen}\multicolumn{7}{c}{\textbf{Proprietary General MLLMs}} \\
    \midrule
    Gemini-3.0-Flash~\cite{google2025gemini3flash} & 64.10 & 0.079 & 38.58 & 55.17 & 87.27 & 0.50 \\
    Gemini-3.0-Pro~\cite{google2025gemini3} & 63.62 & 0.070 & 47.48 & \textbf{82.76} & 89.09 & 0.54 \\
    Gemini-3.1-Pro~\cite{google2026gemini31pro} & 65.63 & \underline{0.068} & 51.34 & \underline{79.31} & 89.09 & 0.56 \\
    Gemini-3.5-Flash~\cite{google2026gemini35flash} & 66.95 & 0.076 & 41.54 & 48.28 & 67.27 & 0.50 \\
    GPT-5.4~\cite{openai2026gpt54} & 64.72 & 0.093 & 40.06 & 10.34 & 85.45 & 0.49 \\
    GPT-5.5~\cite{openai2026gpt55} & 65.44 & 0.091 & 41.84 & 10.34 & 81.82 & 0.48 \\
    \midrule
    \rowcolor{myblue}\multicolumn{7}{c}{\textbf{Specialized Aesthetic Cropping Models}} \\
    \midrule
    CACNet~\cite{hong2021composing} & 68.29 & 0.080 & 54.08 & 0.00 & 0.00 & 0.52 \\
    UNIC~\cite{liu2023beyond} & 62.46 & 0.081 & 31.12 & 0.00 & 0.00 & 0.29 \\
    InstructCrop~\cite{sheng2025instructcrop} & \underline{69.53} & 0.072 & 56.97 & 0.00 & 0.00 & 0.43 \\
    Venus~\cite{du2026venus} & 69.43 & 0.076 & \underline{57.27} & 0.00 & 3.64 & \underline{0.57} \\
    \midrule
    \textbf{ShutterMuse (Ours)} & \textbf{74.30} & \textbf{0.054} & \textbf{70.03} & \textbf{82.76} & 74.55 & \textbf{0.64} \\
    \bottomrule
  \end{tabular}
  \end{adjustbox}
\end{table}
On the photographer-side guidance subset of our CaptureGuide-Bench, we compared ShutterMuse with a comprehensive range of baseline models, including state-of-the-art closed-source models (GPT-5.5, Gemini-3.5-Flash, etc.), open-source general-purpose MLLMs (Kimi-K2.6 \cite{moonshotai2025kimik26}, Qwen3.5-VL \citep{qwen35blog}, etc.), and specialized aesthetic cropping models (Venus \citep{du2026venus}, etc.). Since there is no dedicated MLLM baseline for aesthetic pose recommendation, we also tested general MLLMs for direct COCO-17 keypoint generation. However, without task-specific fine-tuning, they produced no valid renderable pose outputs under our validity check. We therefore compare ShutterMuse with two strong image-editing foundation models, GPT-Image-2 and
Nano-Banana-Pro. We prompt them to recommend human poses that better match the given scene. All prompt templates used for general MLLM baselines, specialized cropping baselines, and image-editing baselines are provided in Appendix~\ref{app:baseline_prompt_templates}.

\paragraph{Photographer-side guidance.}
Table~\ref{tab:photographer_side_bench} shows that ShutterMuse achieves the best overall balance between composition accuracy and decision making on CaptureGuide-Bench. In particular, it attains the highest IoU, lowest BDE, and best refinement success rate. By contrast, specialized models such as InstructCrop and Venus achieve competitive crop quality but perform poorly on \textit{reject} and \textit{keep} decisions. General-purpose MLLMs are better at three-way decision making, but their crop predictions are less accurate.

\paragraph{Subject-side guidance results.}
As shown in Table~\ref{tab:subject_side_bench}, ShutterMuse achieves competitive performance across plausibility, interaction, and aesthetics, obtaining a mean score of 0.34, close to GPT-Image-2 (0.35). The slight advantage of image-editing foundation models is expected, as they benefit from larger model capacity and broad-scale pretraining, which provide strong implicit priors over human anatomy, physical feasibility, spatial interactions, and visual aesthetics. These priors are directly relevant to assessing whether a pose is natural, scene-compatible, and visually pleasing. Notably, ShutterMuse approaches their pose-quality performance while reducing the average inference time by an order of magnitude, making it more suitable for interactive capture-time guidance.

\begin{table}[h]
\centering
\caption{Quantitative results on the subject-side guidance subset of our CaptureGuide-Bench. Mean denotes the average score across Plausibility, Interaction, and Aesthetics. Time and \# Tokens indicate the average time and token usage required to recommend one pose, respectively. Best and second-best results are in \textbf{bold} and \underline{underlined}, respectively.}
\label{tab:subject_side_bench}
\begin{adjustbox}{max width=\linewidth}
\begin{tabular}{lcccccc}
\toprule
\textbf{Method} & Plausibility $\uparrow$ & Interaction $\uparrow$ & Aesthetics $\uparrow$ & Mean $\uparrow$ & Time $\downarrow$ & \# Tokens $\downarrow$ \\
\midrule
Nano-Banana-Pro~\cite{google_nanobananapro} & \textbf{0.63} & \textbf{0.35} & \textbf{0.17} & \textbf{0.39} & \underline{55.16} & \underline{1370} \\
GPT-Image-2~\cite{openai_chatgptimages20} & \underline{0.59} & \underline{0.29} & \underline{0.15} & \underline{0.35} & 102.61 & 1427 \\
\rowcolor{myblue}\textbf{ShutterMuse (Ours)} & 0.58 & 0.27 & 0.14 & 0.34 & \textbf{4.96} & \textbf{412} \\
\bottomrule
\end{tabular}
\end{adjustbox}
\end{table}

\begin{figure}[t]
    \centering
    \includegraphics[width=1.0\linewidth]{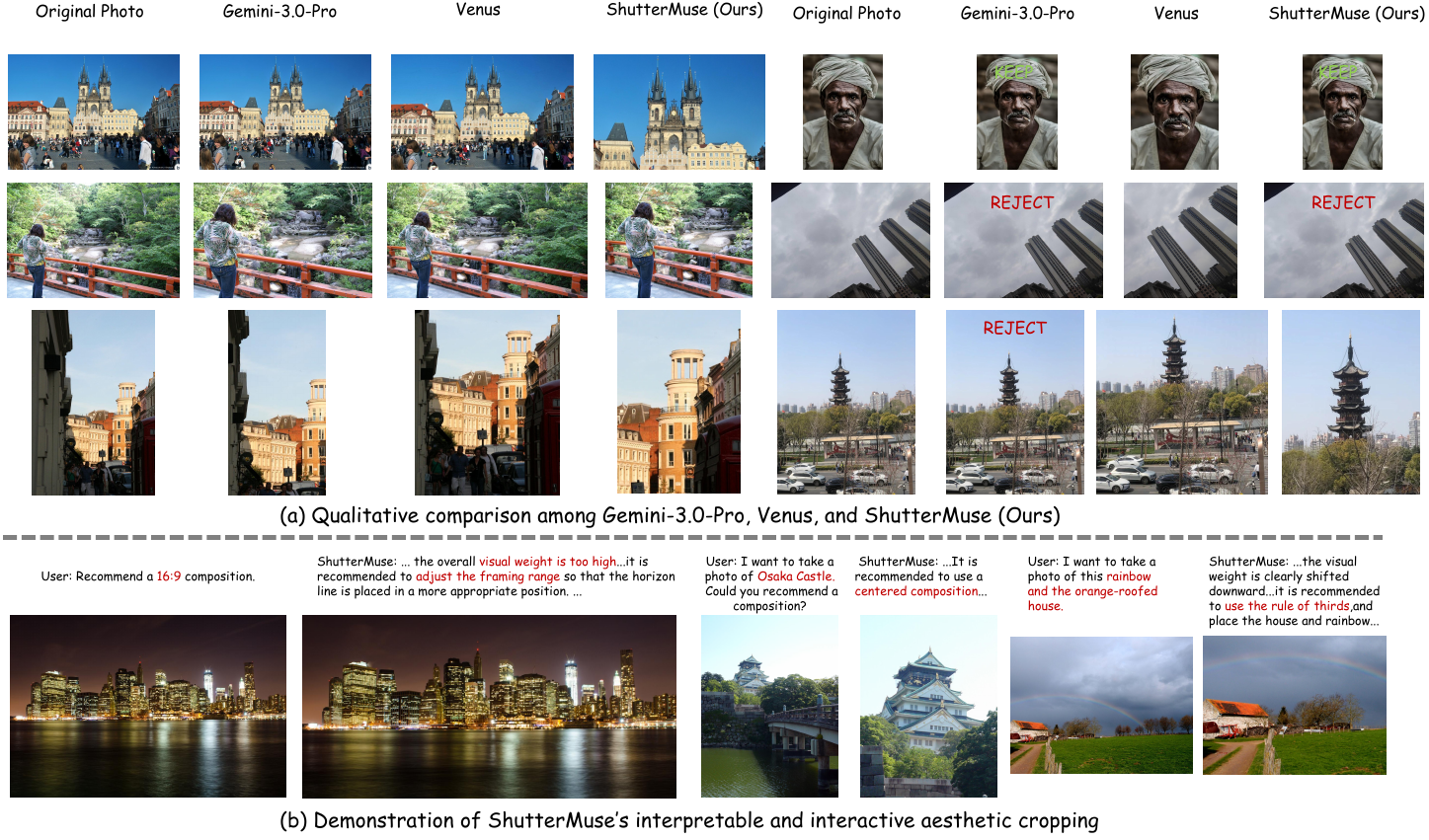}
    \caption{
    Qualitative comparisons of different models on photographer-side guidance.
    }
\label{fig:qualitative_comparison_photo}
\end{figure}
\begin{figure}[t]
    \centering
    \includegraphics[width=1.0\linewidth]{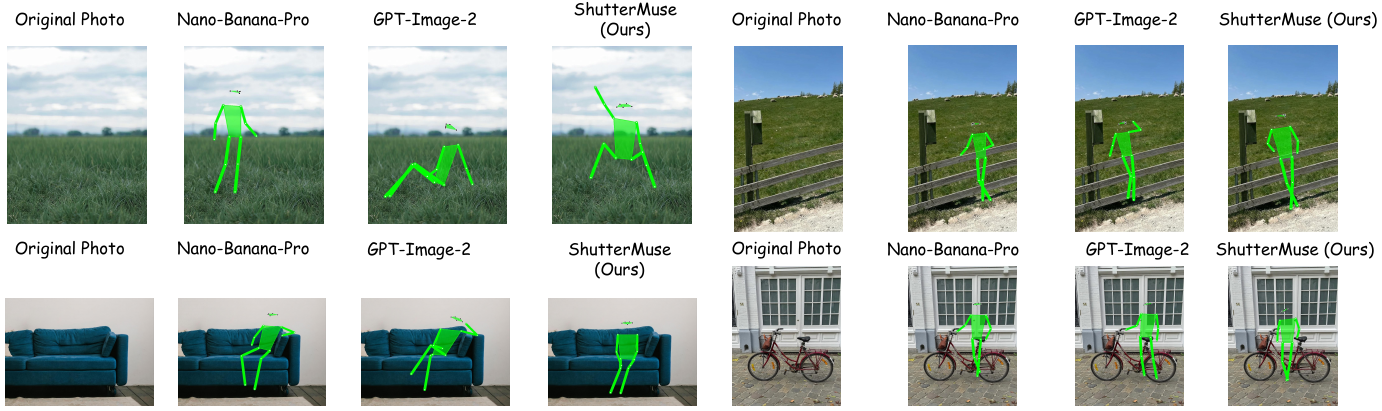}
    \caption{
    Qualitative comparisons of different models on subject-side guidance.
    }
\label{fig:qualitative_comparison_subject}
\end{figure}

\subsection{Qualitative Analysis}

Fig.~\ref{fig:qualitative_comparison_photo} and Fig.~\ref{fig:qualitative_comparison_subject} provide qualitative comparisons for photographer-side and subject-side guidance, respectively. As shown in Fig.~\ref{fig:qualitative_comparison_photo}(a), ShutterMuse makes appropriate guidance decisions and produces aesthetically refined results across diverse capture-time photography scenarios. In contrast, Venus, as a specialized cropping model, tends to optimize composition for all inputs, even when cropping is unnecessary. Fig.~\ref{fig:qualitative_comparison_photo}(b) further shows that ShutterMuse provides interpretable aesthetic rationales that explain its decisions and support interactive refinement, enabling more explainable and personalized aesthetic enhancement. Fig.~\ref{fig:qualitative_comparison_subject} presents qualitative comparisons among Nano-Banana-Pro, GPT-Image-2, and ShutterMuse on the subject-side guidance subset of CaptureGuide-Bench. Across diverse scenes, ShutterMuse generates pose recommendations that are comparable to those produced by image-editing foundation models. Its recommendations are well aligned with the scene context and cover diverse pose types with fine-grained action details.
\subsection{Ablation Study}

\begin{table*}[!t]
\centering
\caption{
Ablation study of training stages and reward components. We compare SFT-only training,
GRPO variants with one reward component removed, and the full ShutterMuse-RL model.
Best or tied-best results are in bold.
}
\label{tab:ablation}
\small
\setlength{\tabcolsep}{5pt}
\renewcommand{\arraystretch}{1.05}
\begin{adjustbox}{max width=\textwidth}
\begin{tabular}{lccccccc}
\toprule
& \multicolumn{4}{c}{Photographer-side Guidance} 
& \multicolumn{3}{c}{Subject-side Guidance} \\
\cmidrule(lr){2-5} \cmidrule(lr){6-8}
Method
& IoU\% $\uparrow$ 
& RSR\% $\uparrow$ 
& KSR\% $\uparrow$ 
& MLLM-Score $\uparrow$ 
& Plausibility $\uparrow$ 
& Interaction $\uparrow$ 
& Aesthetics $\uparrow$ \\
\midrule
ShutterMuse-SFT
& 72.39 & 68.97 & 63.64 & 0.56 & 0.52 & 0.25 & \textbf{0.14} \\
\midrule
ShutterMuse-RL w/o $R_{\mathrm{dec}}$
& 74.10 & 62.07 & 65.45 & 0.62 & 0.56 & \textbf{0.27} & 0.12 \\
ShutterMuse-RL w/o $R_{\mathrm{mask}}$
& 73.76 & 72.41 & 63.63 & 0.61 & 0.54 & \textbf{0.27} & 0.12 \\
ShutterMuse-RL w/o $R_{\mathrm{sub}}$
& 73.49 & 79.31 & 70.91 & \textbf{0.64} & 0.53 & \textbf{0.27} & 0.11 \\
\rowcolor{myblue}
ShutterMuse-RL (Ours)
& \textbf{74.30} & \textbf{82.76} & \textbf{74.55} & \textbf{0.64} 
& \textbf{0.58} & \textbf{0.27} & \textbf{0.14} \\
\bottomrule
\end{tabular}
\end{adjustbox}
\end{table*}

\paragraph{Effect of Training Strategy and Reward Design.}
We ablate both the multi-stage training strategy and the reward design in Tab.~\ref{tab:ablation}.
Compared with ShutterMuse-SFT, the GRPO stage consistently improves photographer-side guidance,
increasing IoU from 72.39\% to 74.30\%, RSR from 68.97\% to 82.76\%, KSR from 63.64\% to
74.55\%, and MLLM-Score from 0.56 to 0.64. It also improves subject-side Plausibility and
Interaction, while Aesthetics remains unchanged. These results suggest that reinforcement
fine-tuning is particularly beneficial for learning discrete capture-time decisions and producing
composition frames aligned with the annotated guidance.
The reward ablations further show that each component contributes to the final performance.
Removing the decision reward substantially degrades RSR and KSR, indicating that
$R_{\mathrm{dec}}$ is critical for learning the \texttt{refine}/\texttt{keep}/\texttt{reject} decision.
Removing the mask reward reduces IoU and MLLM-Score, suggesting that $R_{\mathrm{mask}}$
helps preserve salient regions when predicting refined composition boxes. Removing the
subject-side reward lowers Plausibility, showing that $R_{\mathrm{sub}}$ improves the consistency
between pose recommendations and visible body-part constraints. Overall, the full reward design
achieves the strongest or tied-strongest performance across the evaluated metrics.

\begin{figure}[h]
    \centering
    \includegraphics[width=1.0\linewidth]{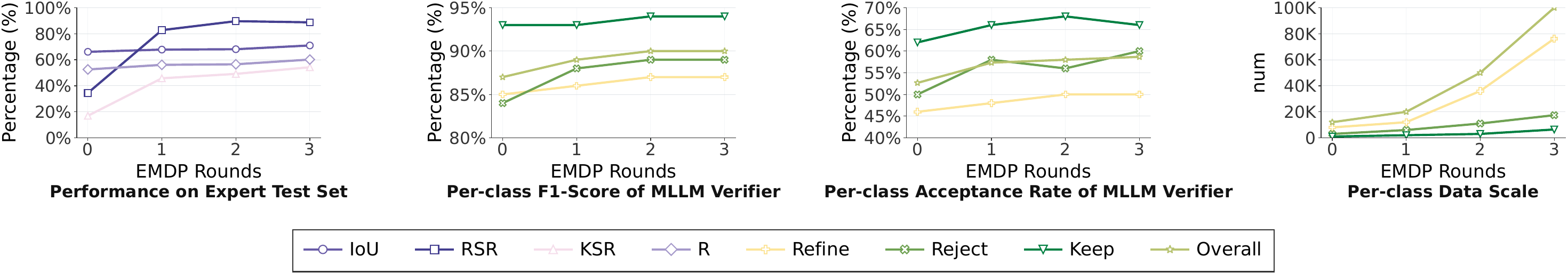}
    \caption{Reliability analysis of EMDP across three rounds. (a) Performance on the expert test set. (b) MLLM Verifier F1-score. (c) Acceptance rate. (d) Training-set size.}
\label{fig:self_distill}
\end{figure}

\begin{wraptable}{r}{0.30\linewidth}
\vspace{-2.0em}
\centering
\caption{
Ranking agreement between MLLM-Score and human preference. Lower ranks indicate better performance.
}
\label{tab:user_study_rankings}
\scriptsize
\setlength{\tabcolsep}{3pt}
\begin{tabular}{@{}lcc@{}}
\toprule
\multicolumn{3}{c}{\textbf{Photographer-side}} \\
\midrule
Method & MLLM & Human \\
\midrule
\rowcolor{myblue}\textbf{ShutterMuse} & \textbf{1} & \textbf{1} \\
Venus & 2 & 2 \\
Gemini-3.0-Pro & 3 & 4 \\
GPT-5.5 & 4 & 3 \\
InstructCrop & 5 & 5 \\
\midrule
\midrule
\multicolumn{3}{c}{\textbf{Subject-side}} \\
\midrule
Method & MLLM & Human \\
Nano-Banana-Pro & 1 & 1 \\
GPT-Image-2 & 2 & 2 \\
\rowcolor{myblue}\textbf{ShutterMuse} & \textbf{3} & \textbf{3} \\
\bottomrule
\end{tabular}
\end{wraptable}
\paragraph{Reliability Analysis of EMDP.}
We analyze the reliability of the proposed expert-seeded, MLLM-verified self-distillation pipeline (EMDP) over three rounds. Before training any composition model, we reserve 450 expert-annotated
samples as a fixed held-out expert test set. These samples are excluded from the initial seed training
set, the EMDP expansion process, SFT, RFT, verifier calibration, and hyperparameter selection. To evaluate the verifier's reliability, we randomly sample 100 examples per category at each round and compare the MLLM verifier's decisions with expert judgments. As shown in Fig.~\ref{fig:self_distill}(a), performance on the held-out expert test set consistently improves, with IoU increasing from 66.11\% to 70.99\%, RSR from 34.48\% to 88.77\%, KSR from 16.95\% to 54.24\%, and R from 52.54\% to 60.15\%. This trend indicates that the progressively accumulated EMDP data effectively improves model performance. 
Fig.~\ref{fig:self_distill}(b) shows that the verifier maintains an F1-score above 87\% across all data categories and rounds, demonstrating the reliability of Gemini-3.0-Pro as the MLLM verifier. Fig.~\ref{fig:self_distill}(c) reports an acceptance rate above 52\%, suggesting that EMDP maintains a stable data generation efficiency. Finally, Fig.~\ref{fig:self_distill}(d) shows that EMDP expands the training set from the initial expert-labeled seed set to 100K samples after three rounds.

\section{User Study}
We further conduct a user study to evaluate the consistency between MLLM-based assessment and human preferences. We randomly sample 100 test examples from each subset of CaptureGuide-Bench and recruit six participants for blind evaluation. Specifically, the photographer-side study compares Gemini-3.0-Pro, GPT-5.5, InstructCrop, Venus, and ShutterMuse, while the subject-side study compares Nano-Banana-Pro, GPT-Image-2, and ShutterMuse. We aggregate the rankings over all examples and participants, and compute Spearman's rank correlation coefficient (SRCC) between the aggregated human ranking and the MLLM-Score ranking on the photographer-side subset. As shown in Table~\ref{tab:user_study_rankings}, the MLLM-based ranking is highly consistent with human preference, achieving an SRCC of $0.90$. On the subject-side subset, the MLLM-based ranking is identical to the aggregated human ranking. These results suggest that the proposed MLLM-based evaluation aligns well with human judgments across both subsets.
\section{Conclusion}

We introduced \textbf{CaptureGuide-Bench}, a benchmark for evaluating MLLMs in capture-time photography guidance, covering both photographer-side composition refinement and subject-side pose recommendation. We constructed \textbf{CaptureGuide-Dataset}, a large-scale dataset with structured annotations and textual rationales, and proposed \textbf{ShutterMuse}, a unified MLLM trained with supervised fine-tuning and reinforcement fine-tuning. Experiments on CaptureGuide-Bench show that ShutterMuse achieves the best overall photographer-side performance among evaluated baselines, especially in balancing refinement accuracy with \textit{keep}/\textit{reject} decisions, and provides competitive subject-side pose recommendations with substantially lower inference cost. This work suggests that MLLMs can serve as practical interactive assistants for photography during capture.


\bibliography{main}
\bibliographystyle{iclr2026_conference}

\appendix

\section{Annotation Protocol}
\label{app:annotation_protocol}
\subsection{Photographer-side Annotation Guidelines}
\label{app:photographer_guidelines}
\begin{figure}[t]
    \centering
    \includegraphics[width=\linewidth]{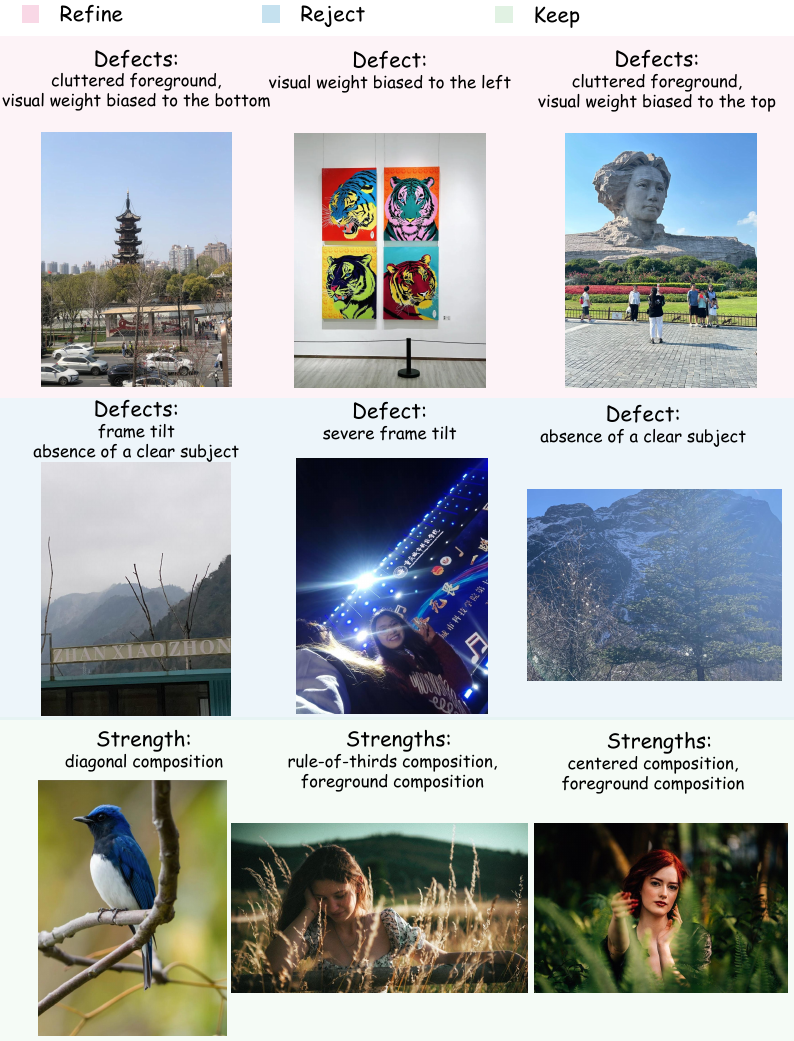}
    \caption{
    Examples from the annotation guidance. The first row shows \texttt{refine} cases, where the image is suitable for recommendation after recomposition; the second row shows \texttt{reject} cases, where non-croppable defects make the image unsuitable; and the third row shows \texttt{keep} cases, where the original framing is already appropriate.
    }
    \label{fig:anno_guide}
\end{figure}
\paragraph{Refine.}
An image is labeled as \texttt{refine} if the main subject or visual content is usable but the current framing can be improved by cropping or recomposition. The shortcomings of \texttt{refine} samples are typically composition-related and can be mitigated by reframing, including cases where the main subject is insufficiently prominent, the visual center is shifted or unbalanced, the subject occupies too small a region, or distracting background elements weaken the composition. For each \texttt{refine} sample, annotators are required to identify the composition issues present in the image and provide a refined framing box that preserves the key subject and important contextual elements while improving the overall composition.
\paragraph{Keep.}
An image is labeled as \texttt{keep} if the current framing is already compositionally appropriate and further cropping would remove useful context or degrade the image quality. For each \texttt{keep} sample, annotators are required to describe the compositional strengths of the current view, such as clear subject placement, balanced visual structure, appropriate use of background context, and coherent scene layout, and explain why the image can be preserved without further reframing.
\paragraph{Reject.}
An image is labeled as \texttt{reject} if the input is unsuitable for recommendation due to severe blur, occlusion, poor exposure, a missing subject, tilted framing, or other defects that cannot be corrected by cropping or recomposition. For each \texttt{reject} sample, annotators are required to describe the non-croppable issues present in the image and explain why they cannot be effectively resolved through reframing.
\subsection{Subject-side Annotation Guidelines}
\label{app:subject_guidelines}

For subject-side annotations, we follow the COCO-17 human keypoint order: nose, left eye, right eye, left ear, right ear, left shoulder, right shoulder, left elbow, right elbow, left wrist, right wrist, left hip, right hip, left knee, right knee, left ankle, and right ankle. The left/right labels are defined from the subject's anatomical perspective.

Each subject-side instance is stored as a structured annotation containing the task type, a natural-language rationale, normalized keypoint coordinates, and visibility labels. Specifically, \texttt{keypoints\_xyn} contains 17 normalized $(x,y)$ coordinates following the COCO-17 order, and \texttt{visibility} contains the corresponding visibility state for each keypoint. We use a three-level visibility scheme: \texttt{1} indicates that the keypoint is inside the image and visible; \texttt{0} indicates that the keypoint is inside the image but not visible, e.g., due to occlusion; and \texttt{-1} indicates that the keypoint is outside the image boundary.

To verify and correct the keypoint annotations, annotators visualize the skeleton defined by the annotated keypoints on the person-erased background image and compare it with the original image. If the visualized pose is clearly inconsistent with the subject's pose in the original image, the sample is discarded. If the pose is geometrically plausible but the visibility label of a keypoint is incorrect, annotators correct the visibility flag according to the above definition.

We further validate the subject-side rationales through a human consistency check. Five professional photographers inspect the rationales and assess whether the described subject-side cues are sufficient to recover the visualized pose. Rationales that do not support a consistent reconstruction of the intended pose are revised or removed from the final annotations.

\section{MLLM-based Evaluation Prompts}
\label{app:evaluation_prompts}
\subsection{Photographer-side MLLM-Score Prompt}
\label{app:photographer_mllm_score_prompt}

We use the following prompt to assess the quality of the photographer-side recomposition results. The MLLM is instructed to act as a strict photography composition reviewer. It receives a single image, where the model-predicted composition frame is overlaid on the original image as a red box, and evaluates only the composition inside the red box.

\begin{promptbox}[Photographer-side MLLM-Score Prompt]
You are a strict photography composition reviewer. You will see a single image containing the original image with the model-predicted composition frame overlaid as a red box. Please evaluate the quality of the composition corresponding to the red box. Output only one JSON object and do not output Markdown.

Reasons why the original image is good: \texttt{<origin\_good>} \\
Reasons why the original image is poor: \texttt{<origin\_bad>}

\textbf{Three-level scoring rules.}
You only need to score the model-predicted composition frame. The score must be one of \texttt{0}, \texttt{0.5}, or \texttt{1}.

\begin{itemize}
    \item \textbf{Score 0: severe composition problem.}
    Assign a score of 0 if the red box contains any of the following problems:
    \begin{enumerate}
        \item The foreground or scene is cluttered, e.g., crowded tourists at a scenic spot, making the image look casual and preventing the subject from standing out.
        \item For portrait composition, check whether the red box cuts through important joints of the main subject, such as the ankle, knee, or neck. If the boundary is merely close to a joint, it should not be considered as cutting the joint. Only when the red boundary passes through the joint should it be regarded as a severe composition problem. Cutting background pedestrians is not considered a severe composition error.
        \item The crop severely violates the original photographic intention. For example, if the original image is intended to capture both the person and the scenery, but the crop contains only the person or only the scenery; or if the original image is intended as a group photo, but the crop contains only a few individuals.
    \end{enumerate}

    \item \textbf{Score 0.5: partially fixes the original problems.}
    Assign a score of 0.5 if the red box fixes only part of the problems described in the ``poor reasons'' of the original image. For example, if the left-biased visual center is corrected but the low visual center remains unresolved, the score should be 0.5. Each listed problem should be checked individually.

    \item \textbf{Score 1: fixes all original problems and improves aesthetics.}
    Assign a score of 1 if the red box fixes all original problems, introduces no obvious common composition errors, and uses an effective composition technique to improve visual balance, rhythm, layering, or spatial relationships.
\end{itemize}

\textbf{Composition knowledge.}
The following common photography composition techniques should be used as references during evaluation:
\begin{itemize}
    \item \textbf{Centered composition:} placing the subject near the image center, suitable for symmetric scenes or cases where the subject should be emphasized.
    \item \textbf{Rule-of-thirds composition:} placing the subject along the thirds lines or near the intersections of a $3\times3$ grid to create visual tension and dynamics.
    \item \textbf{Diagonal composition:} creating diagonal structures in the image to enhance depth and motion.
    \item \textbf{Symmetric composition:} arranging the subject to form symmetry, producing a stable and solemn visual effect.
    \item \textbf{Leading-line composition:} using environmental lines, such as roads, railings, or rivers, to guide the viewer's attention toward the subject.
    \item \textbf{Negative-space composition:} using a plain background or repeated elements to occupy approximately 80\% of the image, while the subject occupies only a small region.
    \item \textbf{Frame-within-a-frame composition:} using mirrors, doors, windows, caves, branches, or other natural frames to enclose the subject.
    \item \textbf{Triangular composition:} using the subject's shape or pose to form a triangle, thereby improving visual stability.
    \item \textbf{Foreground composition:} using a blurred foreground object to increase layering and spatial depth.
    \item \textbf{Long shot:} in portrait photography, a composition where the full body of the subject is included.
    \item \textbf{Medium shot:} in portrait photography, a composition where the subject is included from the mid-calf or mid-thigh upward.
    \item \textbf{Close-up shot:} in portrait photography, a composition where the subject is included from the waist upward, or where the face is shown in close-up.
\end{itemize}

The following common composition errors should be carefully checked:
\begin{itemize}
    \item \textbf{Incomplete subject:} important parts of the key subject are cropped out, e.g., the top of a person's head is cut off. In group photos, all main subjects should be included. For landscape or architectural composition, this does not apply as long as the important parts remain inside the frame; for example, excluding the base of a statue or the bottom of a mountain is not necessarily considered subject incompleteness.
    \item \textbf{Distracting elements:} irrelevant pedestrians, clutter, or objects in the background distract attention from the subject.
    \item \textbf{Tilted image:} the image is noticeably tilted, such as a slanted ground plane or tilted person. A severely tilted image may indicate that no recommendable composition is available.
    \item \textbf{Insufficient visual space:} when the subject's body or gaze has a clear direction, insufficient space is left in that direction. This is common in rule-of-thirds compositions.
    \item \textbf{Unclear subject:} the image is too cluttered to identify a clear subject, or the potential subject is strongly distracted by other visual elements.
\end{itemize}

The output format must strictly be:
\begin{verbatim}
{"score":0 or 0.5 or 1,
 "level":"L0 or L0.5 or L1",
 "reason":"brief explanation",
 "fixed_origin_bad":0 or 1,
 "used_composition_technique":"optional, brief description"}
\end{verbatim}
\end{promptbox}
\subsection{Subject-side MLLM-Score Prompt}
\label{app:subject_mllm_score_prompt}

We use MLLM-based evaluation to assess the quality of subject-side pose recommendations from three complementary dimensions: physical plausibility, scene interaction, and pose aesthetics. Each dimension is evaluated independently using a dedicated prompt. The MLLM receives only one image, i.e., the model-generated pose visualization represented as a skeleton. We normalize the original three-level scores into \(\{0, 0.5, 1.0\}\), where a higher score indicates better quality. We present the prompts below.

\begin{promptbox}[Subject-side MLLM-Score Prompt: Physical Plausibility]
\small

You are a strict but reasonable evaluator for human pose recommendations. This round evaluates only \textbf{physical plausibility}. \\

You will receive only one image: the model-generated pose visualization, shown as a skeleton image. In the skeleton image, red body parts indicate that the model believes these parts are occluded by objects.

There are three types of poses: full-body poses, three-quarter-body poses, and upper-body poses. A full-body pose displays 17 keypoints. For simplicity, the head and the torso are not connected in the visualization. A three-quarter-body pose does not display the left ankle or the right ankle. An upper-body pose does not display the lower body.

\textbf{Scoring rules:}
\begin{itemize}
    \item \textbf{score=0:} The model output severely violates physical constraints, and the pose cannot be imitated by a human. \textbf{Note that the separation between the head and the torso is not a severe physical violation; it is caused by the visualization style. Do not assign score 0 for this reason.}
    \item \textbf{score=0.5:} The pose slightly violates physical plausibility. For example, some red body parts are not actually occluded by objects, or the pose contains minor penetration or floating artifacts. However, the recommended pose can still be understood and imitated.
    \item \textbf{score=1.0:} The body scale is natural with respect to the environment, the contact between the person and the scene is credible, there is no obvious floating or penetration, and the pose can be reasonably imitated by a human.
\end{itemize}

\textbf{Notes:}
\begin{enumerate}
    \item The evaluation does not need to be overly strict; if the pose can be understood and imitated, it is acceptable.
    \item Slight floating is normal for standing poses, because the skeleton visualization only shows keypoints down to the ankles. This should not be regarded as floating.
    \item Crossed legs do not necessarily indicate an error, because the two legs may be positioned at different depths.
    \item When the person is sitting, floating feet can be physically reasonable.
    \item \textbf{Do not mistakenly judge upper-body poses or three-quarter-body poses as incomplete bodies or floating bodies.}
\end{enumerate}

\textbf{Required output format:}
\begin{flushleft}
\ttfamily\footnotesize
\{``score'': 0 or 0.5 or 1.0, ``reason'': ``brief explanation''\}
\end{flushleft}

\end{promptbox}

\begin{promptbox}[Subject-side MLLM-Score Prompt: Scene Interaction]
\small

You are a strict but reasonable evaluator for human pose recommendations. This round evaluates only \textbf{interaction with the scene}.\\

You will receive only one image: the model-generated pose visualization, shown as a skeleton image.

There are three types of poses: full-body poses, three-quarter-body poses, and upper-body poses. A full-body pose displays 17 keypoints. A three-quarter-body pose does not display the left ankle or the right ankle. An upper-body pose does not display the lower body.

\textbf{Scoring rules:}
\begin{itemize}
    \item \textbf{score=0:} Weak interaction. The person does not interact with any object or scene element. Simply standing or squatting on the ground does not count as interaction.
    \item \textbf{score=0.5:} Moderate interaction. The pose shows simple interaction, such as sitting on a chair, leaning against a wall, sitting on steps, or leaning on a railing.
    \item \textbf{score=1.0:} Strong interaction. The pose clearly interacts with distinctive objects or scene elements, including interaction through body movement or gaze. Examples include interacting with small animals, flowers, signs, scenery, or other scene-specific objects.
\end{itemize}

\textbf{Required output format:}
\begin{flushleft}
\ttfamily\footnotesize
\{``score'': 0 or 0.5 or 1.0, ``reason'': ``brief explanation''\}
\end{flushleft}

\end{promptbox}

\begin{promptbox}[Subject-side MLLM-Score Prompt: Pose Aesthetics]
\small

You are a strict but reasonable evaluator for human pose recommendations. This round evaluates only \textbf{pose aesthetics}.\\

You will receive only one image: the model-generated pose visualization, shown as a skeleton image.

There are three types of poses: full-body poses, three-quarter-body poses, and upper-body poses. A full-body pose displays 17 keypoints. A three-quarter-body pose does not display the left ankle or the right ankle. An upper-body pose does not display the lower body.

\textbf{Scoring rules:}
\begin{itemize}
    \item \textbf{score=0:} The pose is an ordinary static pose, with neither dynamic movement nor meaningful pose details, such as standing upright or sitting straight.
    \item \textbf{score=0.5:} The pose contains certain details or dynamics, such as standing with crossed legs, raising one leg, crossing the legs while seated, walking, looking back at the camera, or waving.
    \item \textbf{score=1.0:} The pose contains rich action details, has high visual interest and expressive tension, and is well adapted to the environment.
\end{itemize}

\textbf{Notes:}
Only evaluate the pose itself. Do not consider irrelevant factors such as the person's appearance, clothing, or image composition.

\textbf{Required output format:}
\begin{flushleft}
\ttfamily\footnotesize
\{``score'': 0 or 0.5 or 1.0, ``reason'': ``brief explanation''\}
\end{flushleft}
\end{promptbox}

\section{Baseline Prompt Templates}
\label{app:baseline_prompt_templates}

This section summarizes the prompt templates used for different baseline methods. For specialized cropping baselines, we use the official checkpoints and their corresponding default inference settings. For methods whose original papers provide task-specific prompts, we directly follow the prompts reported in the papers. For general MLLM baselines, we use a unified prompt template to ensure a fair comparison across different models. We present the prompts below.

\subsection{Photographer-side Prompt}
\label{app:photographer_side_baseline_prompt}

\paragraph{Specialized Cropping Baselines.}
For specialized cropping baselines, including InstructCrop and Venus, we use their official checkpoints and default settings. We follow the prompt templates provided in their original papers.

\begin{promptbox}[Venus Prompt]
\small
Please provide the bounding box coordinate of the most visually balanced and aesthetically pleasing composition area.
\end{promptbox}

\begin{promptbox}[InstructCrop Prompt]
\small
Please suggest the best cropping area in this image, and explain the reason.
\end{promptbox}

\paragraph{General MLLM Baselines.}
For general MLLM baselines, we use the following unified prompt. The prompt asks the model to first determine whether the original image has good composition, improvable composition, or extremely poor composition that is difficult to salvage. It then requires the model to output either a normalized bounding box with the specified aspect ratio or the special token \texttt{<non>} when no aesthetically valid crop exists.

\begin{promptbox}[Unified Photographer-side Prompt for General MLLM Baselines]
\small

\textbf{Role.}
You are a senior photography composition mentor.

\vspace{0.5em}
\noindent\textbf{Input.}
You will receive one image, i.e., the original image. Its composition may fall into one of the following three cases, and you need to determine which case it belongs to:
\begin{enumerate}
    \item The original image has excellent composition.
    \item The original image has room for compositional improvement.
    \item The original image has extremely poor composition and is difficult to salvage.
\end{enumerate}

\vspace{0.5em}
\noindent\textbf{Task.}
Based on the composition quality of the original image, generate a logically coherent, natural, and concise analysis. Do not use a list format. The output should be concise and accurate, should follow the provided reference strengths, and should not introduce subjective assumptions or unsupported inferences. Keep the analysis within 100 characters.

The reasoning logic should be as follows.

\vspace{0.5em}
\noindent\textbf{Case 1: The original image has excellent composition.}
If the original image already has excellent composition, first identify the photographic subject. If the subject includes both a person and other elements, such as a person--scenery photo, also specify the scenery included in the subject. Then explain why the composition is aesthetically pleasing, explicitly state that no modification is needed and that preserving the original image is the best choice. Finally, output the bounding box of the entire image.

\vspace{0.5em}
\noindent\textbf{Case 2: The original image has room for compositional improvement.}
If the original image has room for improvement, first determine whether the subject is not prominent. If so, directly state that the subject in the original image is not prominent. Otherwise, identify the photographic subject. If the image is intended as a person--scenery photo, explicitly state that the photographic intention is to capture both the person and the scenery.

Then describe the strengths and weaknesses of the original composition. Finally, if the photographic subject is a person, first examine the original image and determine an appropriate shot scale. A close-up shot refers to a half-body portrait, a long shot refers to a full-body portrait, and a medium shot refers to a portrait cropped from around the knees upward. After specifying the recommended shot scale, explain how to modify the original image to improve its composition, and provide the cropping bounding box.

\vspace{0.5em}
\noindent\textbf{Case 3: The original image has extremely poor composition and is difficult to salvage.}
If the original image has extremely poor composition and is difficult to salvage, first determine whether the subject is not prominent. If so, directly state that the subject in the original image is not prominent. Otherwise, identify the photographic subject. If the image is intended as a person--scenery photo, explicitly state that the photographic intention is to capture both the person and the scenery.

Then point out the defects in the image. If the reference defects include ``cluttered background'' or ``cluttered foreground'', specify what causes the clutter, for example, ``the crowded pedestrians at the bottom cause foreground clutter''. Finally, explicitly state that ``there is no composition scheme in this image that satisfies aesthetic standards'', and output the special token \texttt{<non>}.

\vspace{0.5em}
\noindent\textbf{Bounding box format.}
If a bounding box needs to be returned, it must follow the aspect ratio specified by \texttt{<target\_ratio>} and be returned in the format \texttt{[x1,y1,x2,y2]}, where \texttt{(x1,y1)} is the top-left corner and \texttt{(x2,y2)} is the bottom-right corner. All coordinates must be normalized to the range \([0.0, 1.0]\).

\end{promptbox}

\subsection{Subject-side Prompt}
\label{app:subject_side_baseline_prompt}

For the subject-side task, specialized cropping baselines such as InstructCrop and Venus are not directly applicable, because they are designed to predict cropping bounding boxes rather than generate or recommend human poses. For general MLLM and image-editing baselines, we use a unified editing prompt to request a human mesh that serves as a pose recommendation in the input scene.

To ensure a fair comparison, we do not directly evaluate the edited images produced by different baselines, since their visual quality may be affected by rendering style, texture, lighting, or image-editing artifacts. Instead, for each baseline output, we extract human keypoints from the edited image and re-visualize them into the same skeleton format used by our method. The MLLM-based subject-side evaluation is then performed on these standardized skeleton visualizations. This protocol encourages the evaluator to focus on pose quality, physical plausibility, and scene interaction, rather than the appearance quality of the generated mesh or edited image.

\begin{promptbox}[Unified Subject-side Editing Prompt for General MLLM Baselines]
\small

Please edit this image by adding a human mesh into the scene for photography pose recommendation. The pose of the human mesh must be highly compatible with the current environment, visually aesthetic and natural, and physically plausible.

\vspace{0.5em}
\noindent\textbf{Hard requirements:}
\begin{enumerate}
    \item The pose should be stable, without floating or penetration. The limb proportions should be natural, and the joint directions should be reasonable.
    \item The pose should be aesthetically pleasing and useful as a photography pose reference. Avoid stiff, distorted, or unnatural movements.
    \item The relationship between the human mesh and the environment should be reasonable, e.g., standing, sitting, or leaning on a physically supportive location, with natural body orientation and gaze direction.
    \item Keep the original scene and composition largely unchanged, and do not add any additional people.
    \item Output a single edited image.
\end{enumerate}
\end{promptbox}
\section{Failure Case Analysis for Subject-side Guidance}
\label{app:limitations}

In this section, we analyze representative failure cases of our method on the subject-side guidance task. These cases reveal several limitations of the current pipeline and help clarify the discrepancy between automatic metrics and perceptual quality in some scenarios.

\begin{figure*}[h]
    \centering
    \includegraphics[width=\linewidth]{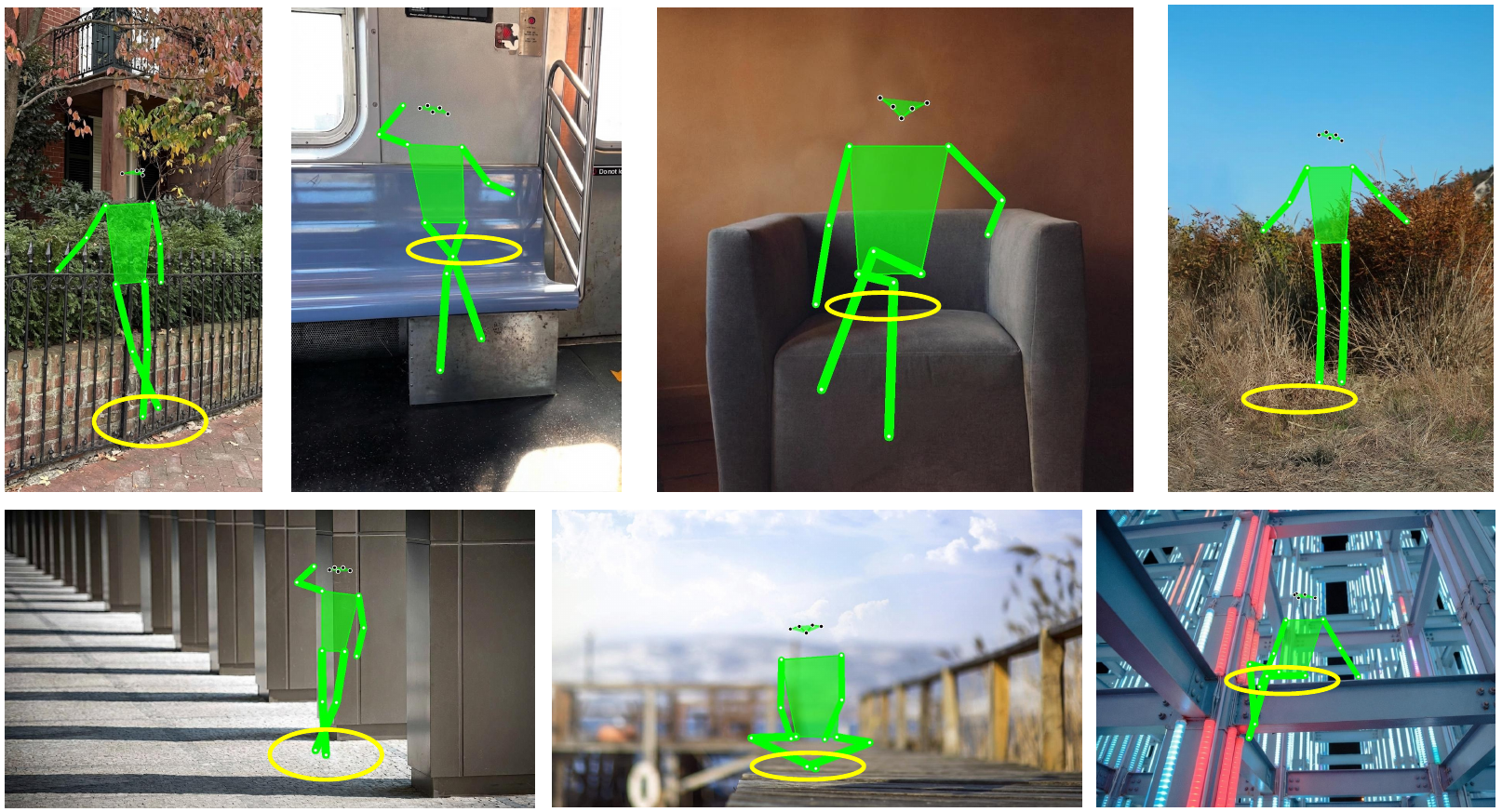}
    \caption{
    Failure cases for subject-side guidance. 
    Some recommended poses appear to float slightly in the skeleton visualization. 
    }
    \label{fig:failure_subject}
\end{figure*}
Figure~\ref{fig:failure_subject} presents representative failure cases for subject-side guidance. The main issue is that the recommended pose may appear slightly floating in the skeleton visualization, especially around the feet. This artifact is largely caused by the 17-keypoint extraction pipeline based on YOLO-style human keypoints. Since the standard 17-keypoint format only localizes the ankles rather than the toes or the full foot contact region, the rendered skeleton may not accurately reflect the actual support area of the human body. As a result, even when the intended pose is physically understandable, the visualization can make the feet appear detached from the ground.

In practice, such slight floating does not substantially affect the usability of the pose recommendation. Users can still understand the intended body configuration and imitate the recommended pose. Nevertheless, this limitation indicates that ankle-only keypoints are insufficient for accurately representing foot--ground contact. Future work could mitigate this issue by incorporating denser body keypoints, explicit foot keypoints, or contact-aware pose representations to better model physical support between the human body and the scene.
\end{document}